\documentclass[11pt]{article}

\usepackage{acl}

\usepackage{times}
\usepackage{latexsym}

\usepackage[T1]{fontenc}

\usepackage[utf8]{inputenc}

\usepackage{silence}
\usepackage{microtype}
\usepackage{inconsolata}

\usepackage{graphicx}

\usepackage{amsmath} 
\usepackage{multirow}
\usepackage{booktabs}
\usepackage{xcolor} 
\usepackage{placeins}
\usepackage{subcaption}
\usepackage{mathtools}
\usepackage{algorithm}
\usepackage{algorithmic}
\usepackage{hyperref}

\makeatletter
\acl@anonymizefalse
\makeatother
\providecommand{\mathbb}[1]{#1}
\providecommand{\blockquote}[1]{#1}

\definecolor{mygreen}{HTML}{008000}
\definecolor{myred}{HTML}{B30000}

\title{Taming the Thinker: Conditional Entropy Shaping for Adaptive LLM Reasoning}

\author{
  \textbf{Shuyu Wei\textsuperscript{1,*}},
  \textbf{Jian Sun\textsuperscript{2,*}},
  \textbf{Delai Qiu\textsuperscript{2}},
  \textbf{Yining Wang\textsuperscript{2}},
  \textbf{Shengping Liu\textsuperscript{2}},
  \textbf{Jiaen Liang\textsuperscript{2}},
\\
  \textbf{Ying Fu\textsuperscript{2}},
  \textbf{Wei Huang\textsuperscript{2}},
  \textbf{Jitao Sang\textsuperscript{1,\ensuremath{\dagger}}}
\\
\\
  \textsuperscript{1}Beijing Key Laboratory of Traffic Data Mining and Embodied Intelligence, \\ 
  Beijing Jiaotong University
\\
  \textsuperscript{2}Unisound AI Technology Co., Ltd.
\\
  \small{
    \textsuperscript{*}Equal contribution. \quad \textsuperscript{\ensuremath{\dagger}}Corresponding author.
  }
}

\begin{document}
\raggedbottom
\maketitle
\begin{abstract}

Entropy-based deep reasoning has emerged as a promising direction for improving the reasoning capabilities of Large Language Models (LLMs), but existing methods often either increase response length indiscriminately or shorten responses at the cost of accuracy. To better balance this trade-off, we introduce \textbf{C}onditional \textbf{E}ntropy \textbf{S}haping (CES), a framework that dynamically controls token-level response entropy, enabling LLMs to produce concise solutions on simple problems while encouraging deeper exploration on hard ones. Built on DAPO, CES uses token-level entropy as an uncertainty signal and applies a conditional bidirectional policy: it penalizes high-entropy ``forking point'' tokens on correct reasoning paths to improve conciseness, and rewards them on incorrect paths to encourage exploration and error correction. We implement CES on DeepSeek-R1-Distill-7B and evaluate it on 12 mathematical benchmarks. CES consistently improves average accuracy while reducing response length relative to DAPO, and supplementary experiments show similar trends on a smaller 1.5B backbone and on out-of-domain benchmarks.

\end{abstract}

\section{Introduction}

In recent years, Large Language Models (LLMs) have demonstrated remarkable capabilities in complex reasoning tasks such as mathematical derivation, code generation, and logical planning \cite{wei2022chain, kojima2022large}. Advanced reasoning models, exemplified by DeepSeek-R1 \cite{guo2025deepseek}, Qwen3 series \cite{yang2025qwen3} and OpenAI o3 series, leverage explicit Chain-of-Thought (CoT) prompting to emulate human-like thought processes, thereby achieving powerful problem-solving abilities. However, the very mechanism that enables this high performance introduces a fundamental tension with a second critical requirement: computational efficiency. The explicit generation of reasoning steps, while crucial for accuracy on complex tasks, inherently increases the number of generated tokens, leading to high latency and computational costs that can hinder real-world applications. This underscores a core dilemma in the field. On one hand, to achieve the highest possible performance, models are encouraged to explore detailed reasoning paths. On the other hand, this may lead to significant inefficiency, a phenomenon often described as ``overthinking'', where models produce unnecessarily lengthy thought processes for trivial questions like ``What is 2+3?'' \cite{chen2024not, ma2025reasoning, yang2025pencil}.

\begin{figure}[t!] 
  \includegraphics[width=\linewidth]{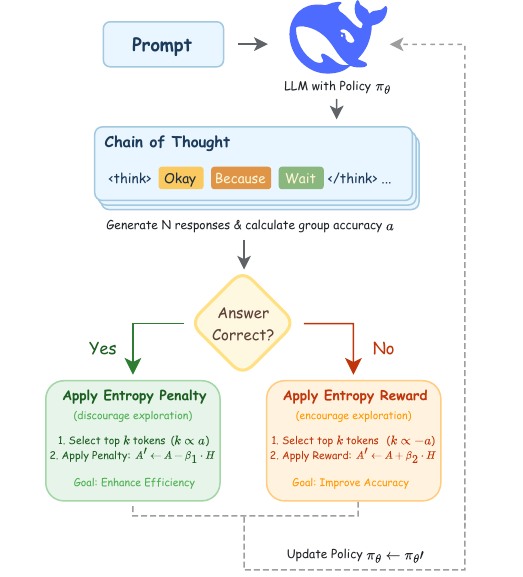}
  \centering
  \caption{Overview of our CES pipeline.} 
  \label{fig:pipeline}
\end{figure}

A novel research direction, which we term entropy-based deep reasoning, has emerged by leveraging token-level entropy to analyze and guide the reasoning process. One study revealed that a few high-entropy tokens within a CoT often act as critical ``forking points'' in the reasoning path, serving as key levers for decision-making \cite{wang2025beyond}. They train the model exclusively on the top 20\% high-entropy tokens and report performance surpassing that of training on all tokens. Another study demonstrated that rewarding high-entropy tokens can encourage model exploration and significantly improve reasoning accuracy \cite{cheng2025reasoning}. Meanwhile, similar work identifies high-covariance tokens as the primary cause of ``entropy collapse'' during reinforcement learning \cite{cui2025entropy}. They restrict their updates to sustain exploration and ultimately improve the model's reasoning accuracy. While these approaches successfully improve model's performance, it comes with the adverse side effect of further elongating the thought process, thereby exacerbating the ``overthinking'' phenomenon and increasing inference costs. 

In parallel, another line of research has focused on improving reasoning efficiency through reinforcement learning, aiming to shorten responses and realize on-demand thinking. Initial efforts included relatively inflexible methods such as post-hoc pruning of generated thoughts \cite{muennighoff2025s1} or training models to adhere to manually specified length budgets \cite{aggarwal2025l1}. More methods have been designed with finer-grained reinforcement learning strategies to achieve the goal of conciseness. For instance, GRPO-LEAD \cite{zhang2025grpo} penalizes correct responses that are longer than average. AdaCoT \cite{lou2025adacot} uses reinforcement learning to learn an optimal policy for triggering the entire CoT process based on query complexity, while Ada-R1 \cite{luo2025ada} first merges long and short CoT models and then uses bi-level preference training to select the most suitable reasoning style for a given problem. While effective at reducing length, these approaches may face a critical trade-off: the gains in efficiency frequently come at the cost of performance degradation on more complex problems that genuinely require deliberate reasoning.

This presents a clear dilemma: methods that enhance efficiency risk may hurt accuracy, while methods that boost accuracy may hurt efficiency. Inspired by recent advances in token-level entropy \cite{wang2025beyond, cheng2025reasoning, cui2025entropy}, our work aims to resolve this trade-off by conditioning the model’s exploratory behavior on the correctness of its reasoning path. In contrast, the previous work \cite{cheng2025reasoning} applies a single, fixed strategy regardless of the reasoning correctness. Based on this core insight, we propose our novel framework \textbf{Conditional Entropy Shaping (CES)}. CES operates within Decoupled Clip and Dynamic Sampling Policy Optimization (DAPO) \cite{yu2025dapo} reinforcement learning framework and intelligently modulates the model's exploratory behavior based on the correctness of its reasoning. As shown in Figure \ref{fig:pipeline}, CES guides the model to:

\begin{enumerate}

\item \textbf{Discourage Exploration}: When a generated reasoning path is \textbf{correct}, CES applies a penalty to the highest-entropy tokens within that path. This encourages the model to become more confident and efficient, refining its thought process toward a concise, direct solution.

\item \textbf{Encourage Exploration}: Conversely, when the path is \textbf{incorrect}, CES rewards these same high-entropy ``forking point'' tokens. This incentivizes the model to explore alternative pathways, and correct its flawed logic.

\end{enumerate}

Empirical results across 12 math benchmarks show that CES improves both accuracy and efficiency on average. The primary contributions of this paper are:

\begin{itemize}

\item We introduce CES, a novel reinforcement learning mechanism that implements a conditional and bidirectional control policy for LLM reasoning.

\item We demonstrate on 12 mathematical benchmarks that CES improves the average accuracy?efficiency trade-off over DAPO. We further show the robustness of CES through additional experiments on smaller 1.5B backbone and out-of-domain benchmarks.

\item We provide a comprehensive analysis of CES's learned behavior, revealing how it develops an adaptive, ``on-demand'' reasoning strategy that strategically allocates computational effort.

\end{itemize}

\section{Method}
Our proposed method, CES, introduces a novel advantage-shaping mechanism into the DAPO framework. DAPO is a reinforcement learning algorithm designed for eliciting complex reasoning in LLMs, which already incorporates several key techniques to stabilize training and improve performance in long CoT scenarios. CES builds upon this strong foundation by introducing an explicit mechanism to manage the trade-off between exploration for accuracy and conciseness for efficiency. It achieves this by dynamically reshaping the token-level advantage signal based on two factors: the correctness of a given model response and the generation entropy of its constituent tokens. Specifically, for correct responses, CES penalizes high-entropy tokens to encourage more direct and concise reasoning paths. Conversely, for incorrect responses, it rewards high-entropy tokens to stimulate exploration and facilitate error correction.

\subsection{Preliminaries: The DAPO Framework}
DAPO enhances the Group Relative Policy Optimization (GRPO) \cite{shao2024deepseekmath} algorithm with a suite of techniques tailored for large-scale reinforcement learning. For a given prompt $x$, a policy $\pi_\theta$ generates a group of $N$ responses, $Y = \{ y_1, y_2, \ldots, y_N \}$. The core of the DAPO objective function is to learn a preference by maximizing the advantage of ``winner'' responses over ``loser'' responses within the group. The full objective is given by:

\begin{equation}
\label{eq:dapo_full}
\begin{aligned}
\mathcal{J}_{\text{DAPO}}(\theta)
={}& \mathbb{E}_{\substack{(q,a)\sim\mathcal{D},\\ \{o_i\}\sim\pi_{\theta_{\text{old}}}}}
\Biggl[
\frac{1}{\sum_{i=1}^{G} |o_i|}
\sum_{i=1}^{G}\sum_{t=1}^{|o_i|} \\
& \min\Bigl(r_{i,t}(\theta)\hat{A}_{i,t}, 
\operatorname{clip}\!\bigl(r_{i,t}(\theta), \\
&\, 1-\epsilon_{\text{low}},\,1+\epsilon_{\text{high}}\bigr)\hat{A}_{i,t}
\Bigr)
\Biggr]
\end{aligned}
\end{equation}

The key components of DAPO relevant to our work are:

\begin{itemize}
    \item  Group-Relative Advantage ($\hat{A}_{i,t}$): The advantage for a response $y_i$ is calculated by normalizing its reward $R_i$ against the mean and standard deviation of rewards within its group $\left\{ R_j \right\}_{j=1}^{G}$. This group-normalized advantage is then applied to every token $t$ in the response $y_i$.
    \item  Token-Level Policy Gradient Loss: DAPO's objective is normalized by the total number of tokens in the batch ($\sum_{i=1}^{G} |o_i|$), ensuring that each token contributes equally to the final loss, regardless of the length of the sequence it belongs to. This prevents shorter sequences from being overshadowed by longer ones.
\end{itemize}

CES intervenes directly at the level of the advantage calculation, $\hat{A}_{i,t}$, before it is used in the DAPO objective function.

\subsection{Conditional Entropy Shaping (CES)}

CES modifies the advantage signal for each token to provide more nuanced guidance to the model. The process involves three steps.

\subsubsection{Step 1: Initial Group-Wise Calculations}

For a given prompt $x$, we generate a response set $Y = \{ y_1, y_2, \ldots, y_N \}$ using the policy $\pi_\theta$. We assign a composite reward $R(y_i)$ to each response, which is the sum of two binary components: an accuracy reward $r_{\text{acc}}(y_i) \in \{0, 1\}$ based on the correctness of the final answer, and a format reward $r_{\text{fmt}}(y_i) \in \{0, 1\}$ for adherence to the \texttt{<think>...</think>} structure. The total reward is $R(y_i) = r_{\text{acc}}(y_i) + r_{\text{fmt}}(y_i)$.

The group accuracy $a$, which is crucial for our conditional mechanism, is computed based only on the correctness reward:
\begin{equation}
a = \frac{1}{N} \sum_{i=1}^{N} r_{\text{acc}}(y_i)
\end{equation}

The initial, unshaped advantage for any token in response $y_i$ is the standard group-normalized advantage, calculated using the total reward $R(y_i)$:
\begin{equation}
A_i = \frac{R(y_i) - \text{mean}(\{R(y_j)\}_{j=1}^N)}{\text{std}(\{R(y_j)\}_{j=1}^N)}
\end{equation}

\subsubsection{Step 2: Dynamic Selection of High-Entropy Tokens}

Then, we compute the token-level entropy. The entropy $H(t_j | y_{i, <j})$ for a token $t_j$ in response $y_i$ at position $j$ is calculated as:

\begin{equation}
H(t_j | y_{i, <j}) = -\sum_{v \in \mathcal{V}} p(v | y_{i, <j}) \log_2 p(v | y_{i, <j})
\end{equation}

In Equation 4, $V$ represents the vocabulary size. We then select the top $k_i$ most entropic tokens in each response $y_i$ to form a set $S_H(y_i)$. The number $k_i$ is determined dynamically to modulate the strength of our intervention:

\begin{equation}
k_i = \lfloor |y_i| \cdot \tau \cdot b_i \rfloor
\end{equation}

In Equation 5, $|y_i|$ is the total length of response $y_i$, and $\tau$ is a base top-rate hyperparameter. The crucial component is the \textbf{dynamic multiplier $b_i$}, defined as:

\begin{equation}
b_i = \begin{cases} a & \text{if } r_{\text{acc}}(y_i) = 1 \\ 1 - a & \text{if } r_{\text{acc}}(y_i) = 0 \end{cases}
\end{equation}

This design aims to apply a stronger intervention (a larger $k_i$) in two specific scenarios: (1) when penalizing a correct response in a group that was easy for the model (high $a$), and (2) when rewarding an incorrect response in a group that was difficult for the model (low $a$).

\subsubsection{Step 3: Entropy-Based Advantage Shaping}

Finally, we compute the reshaped advantage $A'_{i,j}$ for each token $t_j$ in response $y_i$. The advantage is modified only for the selected high-entropy tokens in the set $S_H(y_i)$.

\begin{equation}
A'_{i,j} = 
\begin{cases} 
    A_i - \beta_1 \cdot H(t_j | y_{i, <j}) & \text{if } \substack{r_{\text{acc}}(y_i)=1 \text{ and} \\ t_j \in S_H(y_i)} \\
    A_i + \beta_2 \cdot H(t_j | y_{i, <j}) & \text{if } \substack{r_{\text{acc}}(y_i)=0 \text{ and} \\ t_j \in S_H(y_i)} \\
    A_i & \text{otherwise} 
\end{cases}
\end{equation}

In Equation 7, $\beta_1, \beta_2 > 0$ is a hyperparameter scaling the magnitude of the entropy-based shaping. This final token-level advantage $A'_{i,j}$ replaces the original $\hat{A}_{i,t}$ in the DAPO objective function (Equation 1), thereby injecting our fine-grained control signal into the learning process. The detailed pseudocode for CES is outlined in the Appendix.

\section{Experimental Settings}

\subsection{Backbone Model and Baselines}
Our experiments are conducted in the context of advanced reasoning models. We select the powerful, open-source DeepSeek-R1-Distill-Qwen-7B \cite{guo2025deepseek} as our backbone model, which is known for its strong long-chain reasoning capabilities. To isolate the impact of our proposed method, we establish three baselines for comparison:

\begin{enumerate}

\item \textbf{Original R1-7B}: The pretrained DeepSeek-R1-Distill-Qwen-7B model without any reinforcement learning fine-tuning.

\item \textbf{DAPO Baseline (the key baseline)}: The same backbone model fine-tuned using DAPO algorithm without the CES module. This serves as our primary baseline to directly measure the improvements brought by CES.

\item \textbf{DAPO with ``Entropy Advantage''}: We compare CES with the previous work \cite{cheng2025reasoning}. Their work introduces an ``Entropy Advantage'' that unconditionally adds an entropy-based advantage to all tokens to encourage more exploratory reasoning paths, with the primary goal of improving performance on reasoning tasks. This provides a clear contrast to our conditional, bidirectional approach which aims to balance both accuracy and efficiency.
\end{enumerate}

\begin{table*}[t!]
\centering
\small 
\label{tab:results}
\begin{tabular}{*{11}{c}}
\toprule
\multirow{2}{*}{\textbf{Dataset}} & \multicolumn{2}{c}{\textbf{R1-7B}} & \multicolumn{2}{c}{\textbf{DAPO (Baseline)}} & \multicolumn{2}{c}{\textbf{Entropy Shape}} & \multicolumn{2}{c}{\textbf{CES (Ours)}} & \multicolumn{2}{c}{\textbf{Improvement}} \\
\cmidrule(lr){2-3} \cmidrule(lr){4-5} \cmidrule(lr){6-7} \cmidrule(lr){8-9} \cmidrule(lr){10-11}
& Acc $\uparrow$ & Len $\downarrow$ & Acc $\uparrow$ & Len $\downarrow$ & Acc $\uparrow$ & Len $\downarrow$ & Acc $\uparrow$ & Len $\downarrow$ & Acc $\uparrow$ & Len $\downarrow$ \\
\midrule
AIME24               & 43.3          & 8194 & 42.5          & 7702          & 41.7          & 7527          & \textbf{49.2}      & \textbf{6705} & \textcolor{mygreen}{+6.7} & \textcolor{mygreen}{-997}  \\
AMC23                & 83.8          & 4446 & 85.0          & 4174          & 85.0          & 3843          & \textbf{86.9}      & \textbf{3160} & \textcolor{mygreen}{+1.9} & \textcolor{mygreen}{-1014} \\
CMATH                & 89.2          & 292  & 90.3          & 292           & 90.1          & \textbf{291}  & \textbf{91.9} & 391           & \textcolor{mygreen}{+1.6} & \textcolor{myred}{+99}     \\
CN Middle School 24  & 62.2          & 796  & 61.2          & 815           & 61.4          & \textbf{769}  & \textbf{70.3} & 851           & \textcolor{mygreen}{+9.1} & \textcolor{myred}{+36}     \\
College Math         & 38.8          & 2225 & 39.5          & 1999          & 38.7          & 1850          & \textbf{41.6} & \textbf{1600} & \textcolor{mygreen}{+2.1} & \textcolor{mygreen}{-399}  \\
GaoKao Math Cloze    & 77.1          & 2722 & 77.6          & 2546          & 77.3          & 2538          & \textbf{78.4} & \textbf{1544} & \textcolor{mygreen}{+0.8} & \textcolor{mygreen}{-1002} \\
GaoKao 2023 En       & 71.8          & 2511 & 72.2          & 2182          & 71.8          & 2048          & \textbf{79.9} & \textbf{1914} & \textcolor{mygreen}{+7.7} & \textcolor{mygreen}{-268}  \\
GSM8K                & 86.3          & 443  & 85.5          & \textbf{430}  & 86.5          & 431           & \textbf{89.1} & 462           & \textcolor{mygreen}{+3.6} & \textcolor{myred}{+32}     \\
Minerva Math         & 47.1          & 3063 & \textbf{47.9} & 2546          & 46.7          & 2274          & 42.1               & \textbf{1921} & \textcolor{myred}  {-5.8} & \textcolor{mygreen}{-625}  \\
Olympiad Bench       & 52.3          & 5599 & 53.6          & 5136          & 53.5          & 4821          & \textbf{55.8} & \textbf{4297} & \textcolor{mygreen}{+2.2} & \textcolor{mygreen}{-839}  \\
SVAMP                & 89.4          & 293  & \textbf{90.3} & 290           & \textbf{90.3} & 290           & 90.2               & \textbf{276}  & \textcolor{myred}  {-0.1} & \textcolor{mygreen}{-14}   \\
TABMWP               & 88.0          & 411  & 89.3          & 397           & 88.1          & \textbf{387}  & \textbf{90.0} & 460           & \textcolor{mygreen}{+0.7} & \textcolor{myred}{+63}     \\
\midrule
\textbf{Average}     & 69.1          & 2583 & 69.6          & 2376 & 69.3 & 2256 & \textbf{72.1} & \textbf{1965} & \textcolor{mygreen}{+2.5} & \textcolor{mygreen}{-411}  \\
\bottomrule
\end{tabular}
\caption{Comparison of Accuracy and Response Length on Key Math Datasets. The best result in each category is in \textbf{bold}. The terms ``Acc'' and ``Len'' represent the mean accuracy and the mean response length across 4 assessments for each benchmark.}
\end{table*}

\subsection{Training Details}
We utilize the OpenRLHF framework \cite{hu2024openrlhf} to perform DAPO training, focusing on the domain of solving mathematical problems. Due to resource constraints, our training set only consists of 2500 training samples randomly sampled from the DeepMath dataset \cite{he2025deepmath}. All experiments were carried out on 2 NVIDIA A800 GPUs with 80GB of memory.

Notably, we disable the Dynamic Sampling feature of DAPO when training our CES model. Standard DAPO discards batches where all responses are correct or all are incorrect, as these yield zero advantage and thus no gradient for the sequence-level policy update. However, as CES reshapes advantage at the token level using entropy, these seemingly ``solved'' or ``hopeless'' batches also provide a valuable, non-zero learning signal. This signal is crucial for refining the model's confidence and reasoning style, making every sample useful for training. A comprehensive list of hyperparameters can be found in the Appendix.

\subsection{Evaluation}
For a standardized and reproducible assessment, we employ the evaluation script from the GitHub repository for Qwen2.5-math \cite{Qwen2.5-Math}. To avoid repetition and instability in long-form reasoning models, we adopt a non-greedy decoding strategy, setting a temperature of 0.4, top-$p$ sampling with $p=0.95$, and a repetition penalty of 1.05. For each problem in the test sets, we independently generate 4 responses to ensure a stable and representative measurement. Our evaluation focuses on two primary metrics:

\begin{enumerate}

\item \textbf{Accuracy (Acc)}: The average correctness of the final answers.

\item \textbf{Average Response Length (Len)}: The average number of tokens in the generated responses.

\end{enumerate}

We conduct an extensive evaluation across 12 diverse mathematical reasoning benchmarks: AIME24, AMC23, CMATH, CN Middle School 24, College Math, GaoKao Math Cloze, GaoKao 2023 En, GSM8K, Minerva Math, Olympiad Bench, SVAMP, and TABMWP.

\subsection{Generalization Experiments}
Our main experiments are conducted on DeepSeek-R1-Distill-Qwen-7B. To assess robustness beyond the primary setting, we further replicate CES on a smaller DeepSeek-R1-Distill-1.5B backbone and evaluate the resulting models on out-of-domain coding and general-reasoning benchmarks. These additional results are reported in Appendix E.

\section{Results}
As shown in Table 1, CES demonstrates superior performance by achieving the best overall balance between accuracy and efficiency. On average across all 12 mathematical reasoning datasets, CES achieves the highest accuracy of 72.1\% while simultaneously producing the shortest average response length of 1965 tokens. This represents a significant improvement over our primary baseline, DAPO, with an average accuracy gain of +2.5\% and a substantial average length reduction of 411 tokens.

CES learns to generate more effective and efficient reasoning paths across a wide spectrum of difficulties. For instance, on AIME24, a notoriously difficult competition-level dataset, CES boosts accuracy by a remarkable +6.7\% while cutting the response length by 997 tokens. Similarly, on AMC23 and Olympiad Bench, CES achieves accuracy gains of +1.9\% and +2.2\% respectively, along with massive efficiency improvements, shortening the reasoning paths by 1014 and 839 tokens. This ``win-win'' outcome indicates that CES is not merely pruning the responses, but simultaneously improving the quality and directness of the model's problem-solving strategies. In addition, on test sets such as CN Middle School 24 and GSM8K, it correctly identifies an opportunity where a modest investment in length (+36/+32 tokens) can yield a considerable gain in accuracy (+9.1\%/+3.6\%). This behavior shows that CES is not a naive length reduction algorithm but an intelligent controller that strategically allocates computational budget.

We also observe consistent robustness in the 1.5B backbone. We defer the detailed table to Appendix E.1 due to space limits.

\section{Analysis}

\subsection{Training Dynamics}

\begin{figure*}[t]
    \centering

    \begin{subfigure}[b]{0.32\textwidth}
        \centering
        \includegraphics[width=\textwidth]{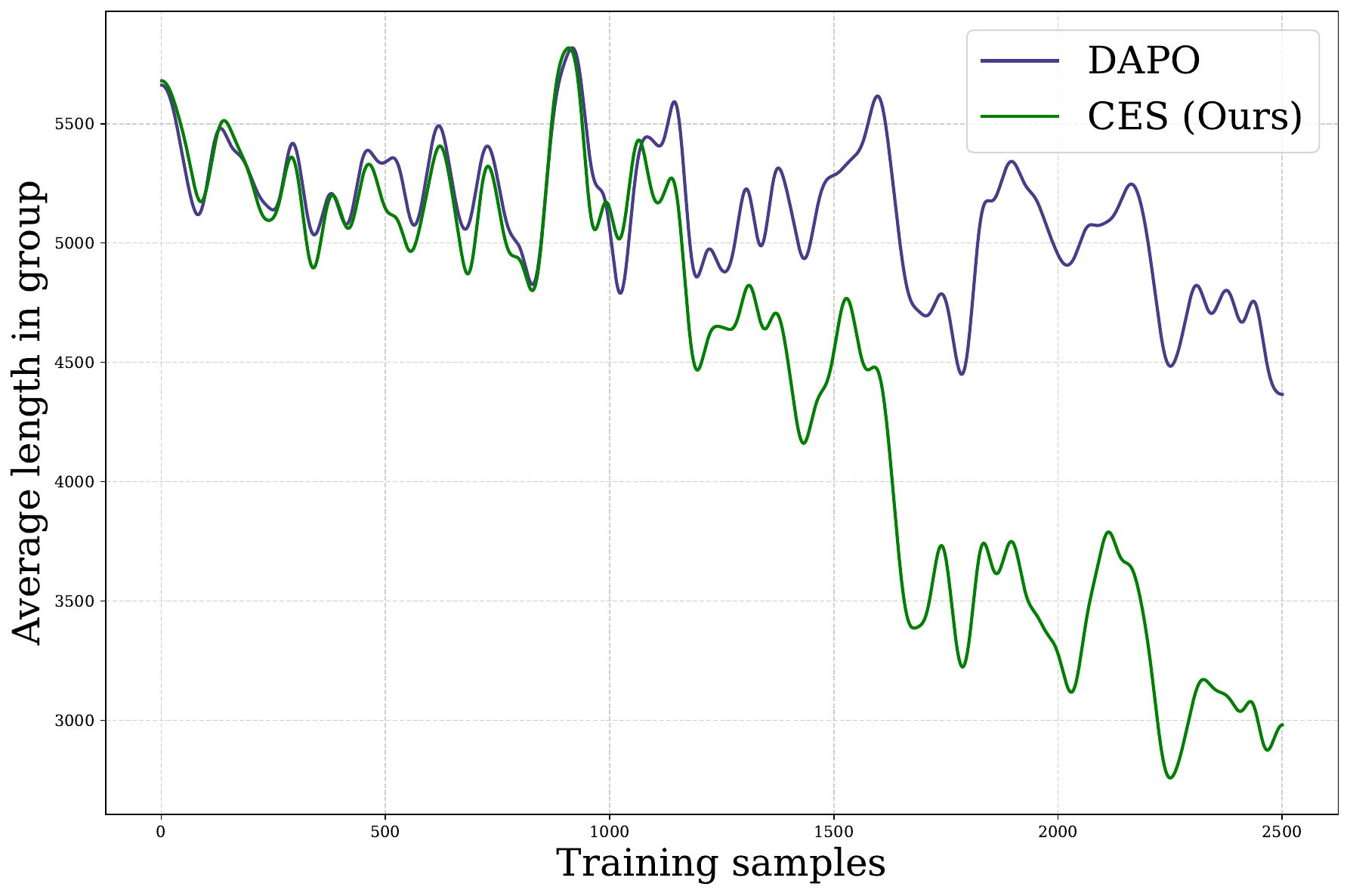} 
        \caption{Response length}
        \label{fig:sub_length}
    \end{subfigure}%
    \hfill%
    \begin{subfigure}[b]{0.32\textwidth}
        \centering
        \includegraphics[width=\textwidth]{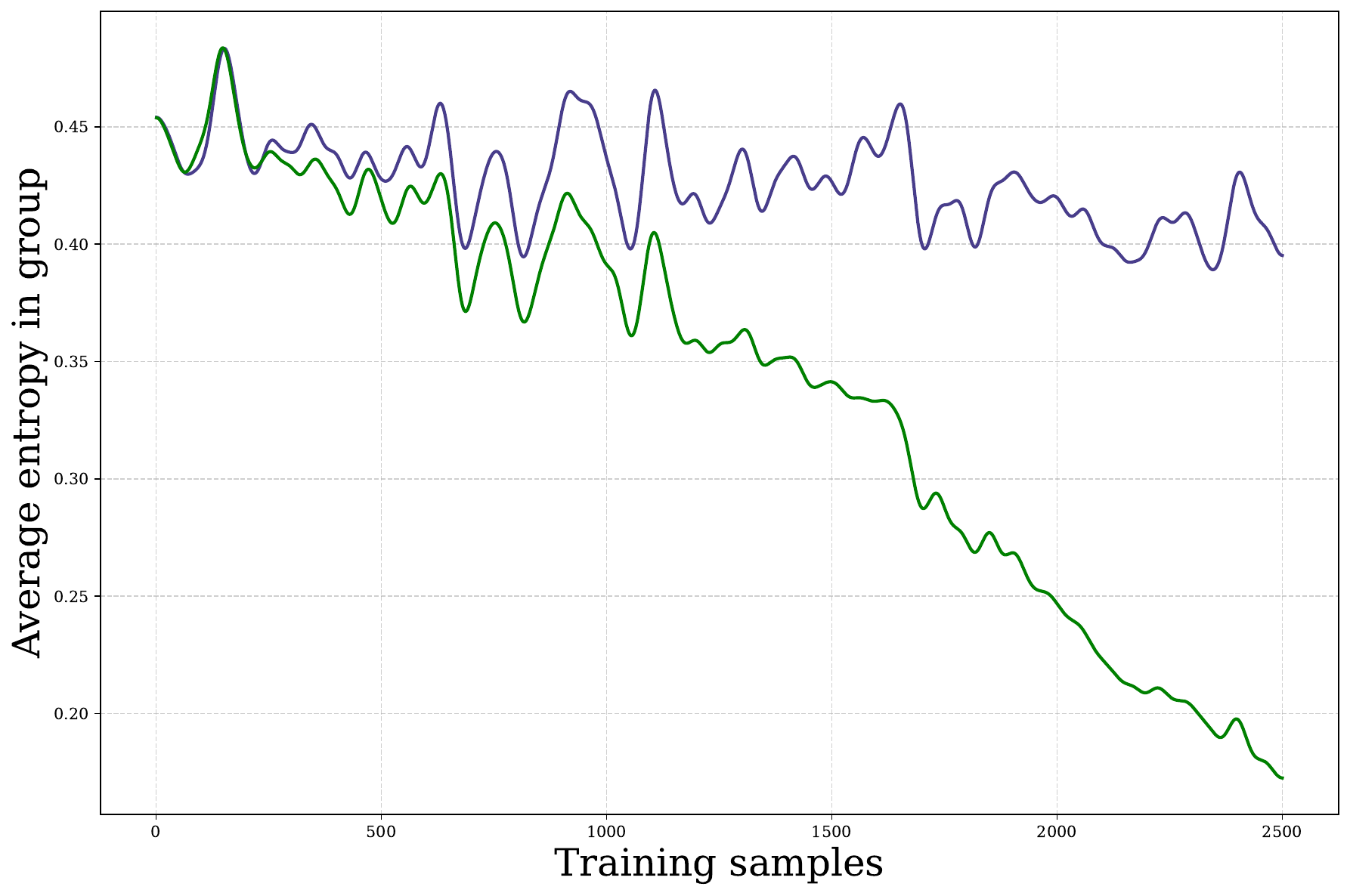}
        \caption{Entropy}
        \label{fig:sub_entropy}
    \end{subfigure}%
    \hfill%
    \begin{subfigure}[b]{0.32\textwidth}
        \centering
        \includegraphics[width=\textwidth]{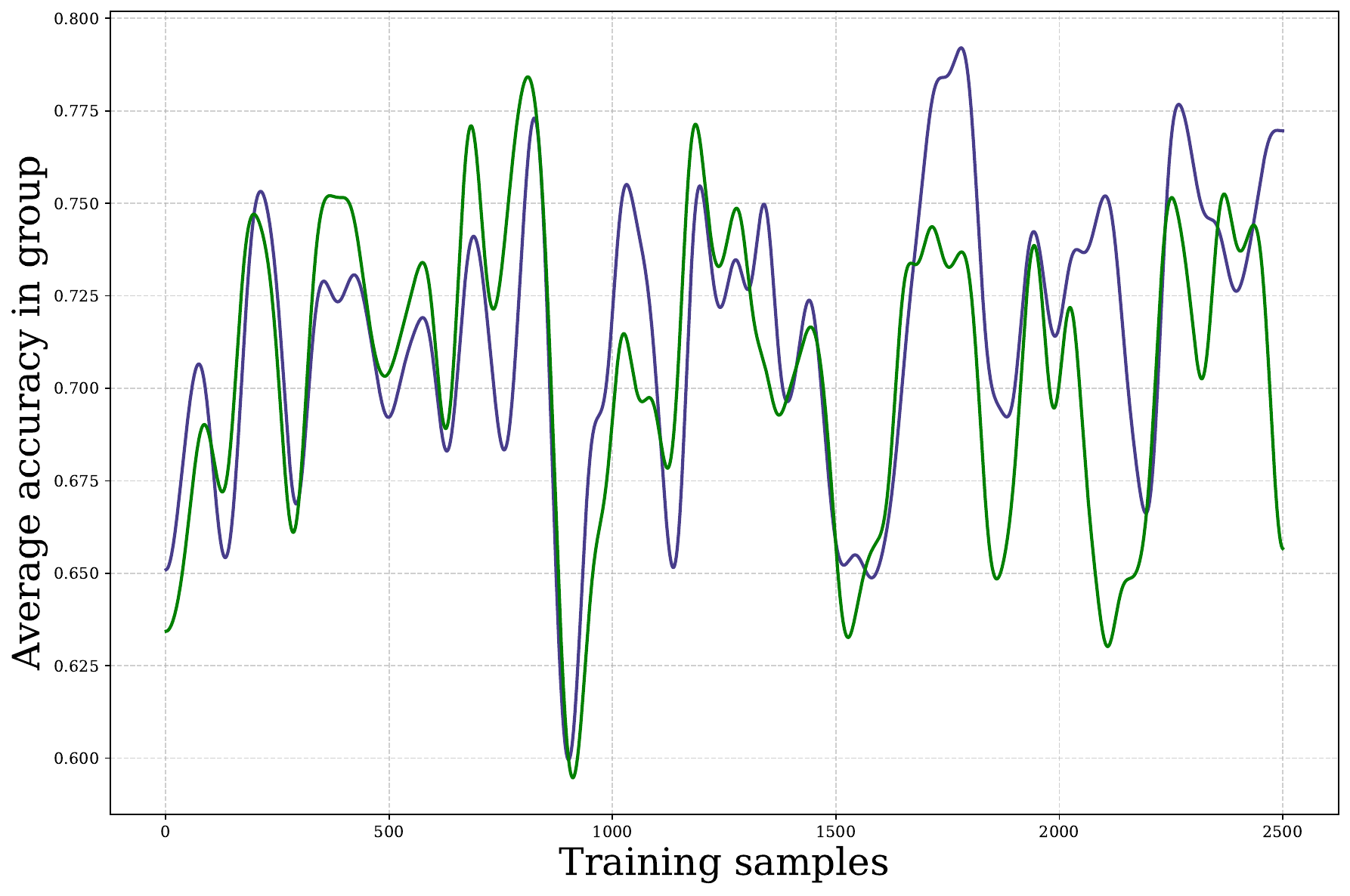}
        \caption{Accuracy}
        \label{fig:sub_accuracy}
    \end{subfigure}

    \caption{Training dynamics of average response length (a), entropy (b), and accuracy (c) for the DAPO baseline (blue) and our CES method (green).}
    \label{fig:training_dynamics_combined}
\end{figure*}

To gain deeper insight into the mechanism of CES, we analyze the evolution of key metrics throughout the training process. Figure \ref{fig:training_dynamics_combined} plots the average response length, average token entropy and average group accuracy, comparing our CES-enhanced DAPO training against the standard DAPO baseline.

A striking pattern emerges in the Response Length and Entropy plots. For the first 1000 training samples, both the CES and baseline models exhibit similar behavior, maintaining a high and stable average length and entropy. This initial phase can be interpreted as the primary task acquisition stage, where both models are focused on learning the fundamental mechanics of solving the problems to achieve a reward. During this period, the policy is highly exploratory, and the CES mechanism has not yet become a dominant optimization force.

However, a clear divergence occurs after 1000 training samples. While the DAPO baseline's length and entropy remain high and relatively constant, the CES model's metrics begin a steep and consistent decline. The average response length drops from over 5000 to nearly 3000 tokens, and the average entropy falls from 0.4 to below 0.2. This second phase demonstrates onset of CES’s core effect, where the entropy penalty on correct answers becomes a powerful and consistent training signal. The model learns that it can maximize its reward not just by being correct, but by being correct and confident. The strong correlation between the decline in entropy and length empirically validates our hypothesis that penalizing high-entropy ``forking points'' effectively prunes unnecessary, verbose exploration, leading to more concise reasoning paths.

In Figure \ref{fig:training_dynamics_combined}(c), we observe that neither the baseline nor the CES model shows a significant, sustained upward trend in accuracy, with both curves fluctuating in a similar trend throughout training. This behavior is likely attributable to the limited size of our training set (2500 samples) and the absence of carefully-designed data strategy. However, it also reflects that the improvements in efficiency (i.e., shorter length and lower entropy) achieved by CES are realized without sacrificing model's problem-solving performance. The CES model maintains an accuracy level competitive with the baseline, while operating at a significantly lower computational budget. In general, these dynamics reveal that CES successfully introduces a distinct optimization phase into training: after the initial task acquisition, it effectively teaches the model to become more efficient and decisive, achieving conciseness without compromising its learned reasoning capabilities.

\subsection{Analysis of increasing response length on simple test sets}

A notable observation from our main results is that while CES significantly shortens responses on most datasets, it increases the average response length on four specific datasets: CMATH, CN Middle School 24, GSM8K and TABMWP. A common characteristic of these datasets is their relatively shorter response length (typically under 1000 tokens) and higher performance, suggesting they are simpler overall. This phenomenon seems to contradict our goal of improving efficiency.

We assume that this is a characteristic of adaptive reasoning manifested by CES. The key to understanding this lies in moving beyond dataset-level averages and analyzing model behavior on a finer-grained, per-question difficulty level. To test this, we stratified the questions within these four datasets into two categories based on the original R1-7B model's performance:

\begin{enumerate}

\item ``Simple Questions'': Questions where the R1-7B model's accuracy is greater than 50\%.

\item ``Difficult Questions'': Questions where the R1-7B model's accuracy is less than or equal to 50\%.

\end{enumerate}

\begin{figure}[t]
    \centering
    \includegraphics[width=\columnwidth]{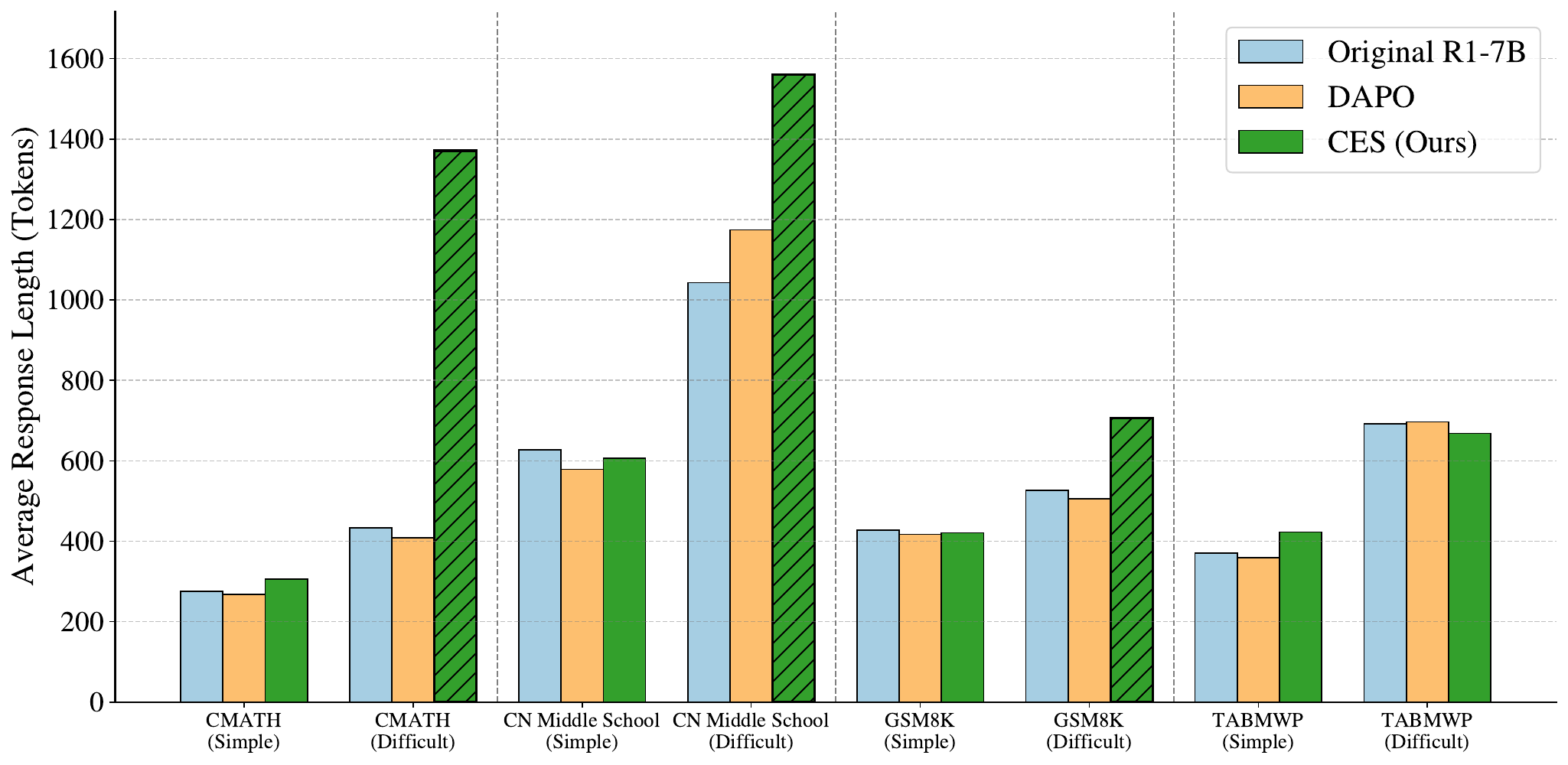}
    \caption{
        Comparison of average response length, stratified by question difficulty on four simpler datasets.
    }
    \label{fig:stratified_length}
\end{figure}

Figure \ref{fig:stratified_length} presents the average response length of both DAPO and CES models across these stratified difficulties. For ``Difficult Questions'' within these simpler datasets, CES triggers a significant increase in response length in most datasets. Conversely, for ``Simple Questions'', the response lengths remain relatively stable or change minimally.

The failure mode of the original R1-7B model on these ``difficult'' questions may be insufficient exploration. Accustomed to the simple patterns of the dataset, it applies a short, inadequate template and fails. CES, through its mechanism of rewarding entropy on incorrect answers, correctly identifies these failures and provides a strong incentive for deeper exploration. It forces the model to abandon the failed template and invest more effort in finding a correct solution. In contrast, on complex datasets like Olympiad Bench, R1-7B's failure mode is often inefficient overthinking, producing long, verbose, and incorrect reasoning. There, CES's primary role is to prune this redundancy. In summary, the strategic investment in reasoning for difficult problems outweighs the minor length changes on simple ones, leading to an increase in the dataset's overall average response length.

\subsection{Ablation Studies}
To validate the key components of our CES framework, we conduct two main ablation studies. These experiments are designed to investigate the importance of our dynamic token selection mechanism and the role of the entropy gradient in our advantage shaping formula.

\begin{table}[ht]
\centering
\begin{tabular}{l c c}
\toprule
\textbf{Method} & \textbf{Acc ↑} & \textbf{Len ↓} \\
\midrule
Original R1-7B & 69.1 & 2583 \\
DAPO (Baseline) & 69.6 & 2376 \\
\midrule
CES w/o Dynamic $b$ & 69.5 & 2462 \\
CES w/o Entropy Gradient & 69.4 & 2539 \\
\midrule
\textbf{CES (Ours)} & \textbf{72.1} & \textbf{1965} \\
\bottomrule
\end{tabular}
\caption{Ablation study on the core components of CES.}
\label{tab:ablation}
\end{table}

\subsubsection{The Importance of Dynamic Token Selection}
A core feature of CES is the dynamic calculation of $k$, the number of high-entropy tokens to be shaped in each response. This number is modulated by a dynamic multiplier $b$ (where $b=a$ for correct responses and $b=1-a$ for incorrect ones, with $a$ being the group accuracy), which adjusts the intervention strength based on the perceived difficulty of the problem. To test the necessity of this design, we trained an ablated model, ``RemoveAcc'', where we removed this dynamic multiplier by fixing $b=1$. In this setting, a constant percentage of tokens with the highest entropy is always selected for entropy shaping, regardless of group accuracy.

The results shown in Table \ref{tab:ablation} indicates that the ``RemoveAcc'' model's average accuracy drops to 69.5\%, nearly identical to the DAPO baseline (69.6\%) and significantly underperforming the full CES model (72.1\%). Furthermore, its average response length increases to 2462, making it even less efficient than the DAPO baseline (2376).

While the behavior of a fixed $b=1$ is identical to our dynamic $b$ at the absolute extremes (when group accuracy $a=1$ or $a=0$), the critical difference emerges in the vast majority of training scenarios where the model's performance is mixed ($0<a<1$). Consider a difficult problem where the model finds a correct solution for the first time, resulting in a low group accuracy (e.g., $a=0.25$). Our full CES method applies a very gentle penalty, scaling the intervention by $b=a=0.25$. This protects the newfound, likely inefficient reasoning path, acknowledging that it is a valuable success on a difficult problem. The ``RemoveAcc'' ablation, in contrast, applies the maximal penalty ($b=1$). It aggressively punishes the high-entropy tokens in this fragile, correct solution, effectively signaling to the model that this ``messy'' path to success is undesirable. This can cause the model to discard the correct reasoning logic in subsequent updates, leading to performance degradation.

Therefore, the dynamic multiplier $b$ acts as a crucial adaptive regularizer. It provides a proportional response: applying gentle, protective pressure on novel solutions to difficult problems, while applying strong, optimizing pressure on mastered solutions to easy problems. By removing this calibrated intelligence, the ``RemoveAcc'' model fails, demonstrating that the dynamic selection of tokens is essential for robustly learning to be both accurate and efficient.

\subsubsection{The Role of the Entropy Gradient in Bidirectional Control}

In our CES formulation, the entropy term $H$ is included in the computation graph, meaning the model's policy is explicitly optimized to produce outputs that align with our entropy-based objectives. However, a related work \cite{cheng2025reasoning} that also uses an entropy-based advantage term introduces a ``detach'' operation in their implementation. This prevents the gradient of the entropy term from being computed, using it only to scale the magnitude of the existing policy gradient rather than setting a new optimization goal. To investigate this choice, we trained an ablated model, where we detached our entropy shaping term from the computation graph.

The results shown in Table \ref{tab:ablation} indicate that this change is detrimental to our method. The Detach model's performance (69.4\%) regresses to that of the DAPO baseline (69.6\%) in accuracy, while its average length balloons to 2539, becoming the least efficient of all training configurations. The reason for this failure lies in the fundamental difference in goals between CES and the method of related work \cite{cheng2025reasoning}. As their objective is unconditional exploration, detaching the entropy term serves as a clever way to amplify the existing policy updates at uncertain steps without asking the model to learn to be ``more uncertain''.

However, CES has a dual, conditional objective. The ``inhibit exploration'' part of our mechanism ($A' \leftarrow A - \beta \cdot H$ for correct answers) is predicated on teaching the model to become more efficient by producing lower-entropy outputs. This requires a non-zero gradient so the model can learn to directly reduce entropy to avoid the penalty. Detaching the term completely breaks this crucial learning signal. Without the gradient, the penalty becomes a simple, static reduction in advantage that provides no direction for how to improve efficiency. This lead to the observed outcome: baseline accuracy with uncontrolled, verbose responses. Therefore, maintaining the entropy gradient is essential for the bidirectional control at the heart of CES to function as intended.

\section{Related Work}

\subsection{Reinforcement Learning for LLMs} 

Reinforcement learning is a core technique for aligning pretrained language models. Early RLHF pipelines commonly relied on Proximal Policy Optimization (PPO) \cite{schulman2017proximal} with a separately trained reward model, while more recent work has shifted toward direct optimization methods to improve stability and simplify training. A representative example is Direct Preference Optimization (DPO) \cite{rafailov2023direct}, which derives an optimization signal directly from preference data. This paradigm has been extended to reasoning settings that compare multiple responses to the same prompt, leading to algorithms such as GRPO and our baseline DAPO, which optimize policies using sequence-level preferences. Building on this line, our method CES introduces a more fine-grained mechanism by intervening at the token level and dynamically shaping the learning signal within the DAPO framework.

\subsection{Entropy in LLMs}

Entropy quantifies the uncertainty of a probability distribution. In LLMs, token-level entropy measures the uncertainty of the predicted distribution over the vocabulary at each generation step: higher entropy corresponds to a flatter distribution and lower confidence in selecting the next token \cite{li2025entropy}.

\section{Conclusion}

In this work, we address the fundamental challenge of balancing performance and efficiency in LLM reasoning. To resolve this trade-off, we propose CES, a framework that enables models to adapt their reasoning strategy: thinking concisely when confident, and reasoning deeply when uncertain. CES achieves consistent improvements in both accuracy and computational efficiency across diverse mathematical reasoning benchmarks, alleviating the inherent trade-off between exploration and exploitation.

Beyond empirical gains, this work suggests a broader principle: LLMs can learn not just to reason accurately, but to regulate how they reason. This opens directions for building fine-grained, resource-aware reasoning systems that require cost-sensitive inference.

\section*{Limitations}

While CES achieves an average win?win by conditionally shaping token-level advantages with entropy, several limitations remain. First, the current formulation still relies on outcome-verifiable correctness signals to compute group accuracy $a$ and to determine the direction of entropy shaping. As a result, applying the same mechanism to tasks with ambiguous, subjective, or weakly verifiable outcomes is less straightforward. Meanwhile, CES remains moderately sensitive to hyperparameters such as $\tau$ and $\beta$. In practice, the method is robust within a reasonable range, but achieving the best accuracy--efficiency balance may still require light calibration when transferring to a new backbone or task distribution.



\bibliography{custom}

@article{wei2022chain,
  title={Chain-of-thought prompting elicits reasoning in large language models},
  author={Wei, Jason and Wang, Xuezhi and Schuurmans, Dale and Bosma, Maarten and Xia, Fei and Chi, Ed and Le, Quoc V and Zhou, Denny and others},
  journal={Advances in neural information processing systems},
  volume={35},
  pages={24824--24837},
  year={2022}
}

@article{kojima2022large,
  title={Large language models are zero-shot reasoners},
  author={Kojima, Takeshi and Gu, Shixiang Shane and Reid, Machel and Matsuo, Yutaka and Iwasawa, Yusuke},
  journal={Advances in neural information processing systems},
  volume={35},
  pages={22199--22213},
  year={2022}
}

@article{guo2025deepseek,
  title={Deepseek-r1: Incentivizing reasoning capability in llms via reinforcement learning},
  author={Guo, Daya and Yang, Dejian and Zhang, Haowei and Song, Junxiao and Zhang, Ruoyu and Xu, Runxin and Zhu, Qihao and Ma, Shirong and Wang, Peiyi and Bi, Xiao and others},
  journal={arXiv preprint arXiv:2501.12948},
  year={2025}
}

@article{yang2025qwen3,
  title={Qwen3 technical report},
  author={Yang, An and Li, Anfeng and Yang, Baosong and Zhang, Beichen and Hui, Binyuan and Zheng, Bo and Yu, Bowen and Gao, Chang and Huang, Chengen and Lv, Chenxu and others},
  journal={arXiv preprint arXiv:2505.09388},
  year={2025}
}

@article{chen2024not,
  title={Do not think that much for 2+3=? on the overthinking of o1-like llms},
  author={Chen, Xingyu and Xu, Jiahao and Liang, Tian and He, Zhiwei and Pang, Jianhui and Yu, Dian and Song, Linfeng and Liu, Qiuzhi and Zhou, Mengfei and Zhang, Zhuosheng and others},
  journal={arXiv preprint arXiv:2412.21187},
  year={2024}
}

@article{ma2025reasoning,
  title={Reasoning models can be effective without thinking},
  author={Ma, Wenjie and He, Jingxuan and Snell, Charlie and Griggs, Tyler and Min, Sewon and Zaharia, Matei},
  journal={arXiv preprint arXiv:2504.09858},
  year={2025}
}

@article{yang2025pencil,
  title={Pencil: Long thoughts with short memory},
  author={Yang, Chenxiao and Srebro, Nathan and McAllester, David and Li, Zhiyuan},
  journal={arXiv preprint arXiv:2503.14337},
  year={2025}
}

@article{wang2025beyond,
  title={Beyond the 80/20 rule: High-entropy minority tokens drive effective reinforcement learning for llm reasoning},
  author={Wang, Shenzhi and Yu, Le and Gao, Chang and Zheng, Chujie and Liu, Shixuan and Lu, Rui and Dang, Kai and Chen, Xionghui and Yang, Jianxin and Zhang, Zhenru and others},
  journal={arXiv preprint arXiv:2506.01939},
  year={2025}
}

@article{cheng2025reasoning,
  title={Reasoning with Exploration: An Entropy Perspective},
  author={Cheng, Daixuan and Huang, Shaohan and Zhu, Xuekai and Dai, Bo and Zhao, Wayne Xin and Zhang, Zhenliang and Wei, Furu},
  journal={arXiv preprint arXiv:2506.14758},
  year={2025}
}

@article{yu2025dapo,
  title={Dapo: An open-source llm reinforcement learning system at scale},
  author={Yu, Qiying and Zhang, Zheng and Zhu, Ruofei and Yuan, Yufeng and Zuo, Xiaochen and Yue, Yu and Dai, Weinan and Fan, Tiantian and Liu, Gaohong and Liu, Lingjun and others},
  journal={arXiv preprint arXiv:2503.14476},
  year={2025}
}

@article{li2025entropy,
  title={Entropy-Aware Branching for Improved Mathematical Reasoning},
  author={Li, Xianzhi and Callanan, Ethan and Zhu, Xiaodan and Sibue, Mathieu and Papadimitriou, Antony and Mahfouz, Mahmoud and Ma, Zhiqiang and Liu, Xiaomo},
  journal={arXiv preprint arXiv:2503.21961},
  year={2025}
}

@article{muennighoff2025s1,
  title={s1: Simple test-time scaling},
  author={Muennighoff, Niklas and Yang, Zitong and Shi, Weijia and Li, Xiang Lisa and Fei-Fei, Li and Hajishirzi, Hannaneh and Zettlemoyer, Luke and Liang, Percy and Cand{\`e}s, Emmanuel and Hashimoto, Tatsunori},
  journal={arXiv preprint arXiv:2501.19393},
  year={2025}
}

@article{zhang2025grpo,
  title={Grpo-lead: A difficulty-aware reinforcement learning approach for concise mathematical reasoning in language models},
  author={Zhang, Jixiao and Zuo, Chunsheng},
  journal={arXiv preprint arXiv:2504.09696},
  year={2025}
}

@article{aggarwal2025l1,
  title={L1: Controlling how long a reasoning model thinks with reinforcement learning},
  author={Aggarwal, Pranjal and Welleck, Sean},
  journal={arXiv preprint arXiv:2503.04697},
  year={2025}
}

@article{schulman2017proximal,
  title={Proximal policy optimization algorithms},
  author={Schulman, John and Wolski, Filip and Dhariwal, Prafulla and Radford, Alec and Klimov, Oleg},
  journal={arXiv preprint arXiv:1707.06347},
  year={2017}
}

@article{rafailov2023direct,
  title={Direct preference optimization: Your language model is secretly a reward model},
  author={Rafailov, Rafael and Sharma, Archit and Mitchell, Eric and Manning, Christopher D and Ermon, Stefano and Finn, Chelsea},
  journal={Advances in neural information processing systems},
  volume={36},
  pages={53728--53741},
  year={2023}
}

@article{shao2024deepseekmath,
  title={Deepseekmath: Pushing the limits of mathematical reasoning in open language models},
  author={Shao, Zhihong and Wang, Peiyi and Zhu, Qihao and Xu, Runxin and Song, Junxiao and Bi, Xiao and Zhang, Haowei and Zhang, Mingchuan and Li, YK and Wu, Yang and others},
  journal={arXiv preprint arXiv:2402.03300},
  year={2024}
}

@article{hu2024openrlhf,
  title={Openrlhf: An easy-to-use, scalable and high-performance rlhf framework},
  author={Hu, Jian and Wu, Xibin and Zhu, Zilin and Wang, Weixun and Zhang, Dehao and Cao, Yu and others},
  journal={arXiv preprint arXiv:2405.11143},
  year={2024}
}

@article{he2025deepmath,
  title={Deepmath-103k: A large-scale, challenging, decontaminated, and verifiable mathematical dataset for advancing reasoning},
  author={He, Zhiwei and Liang, Tian and Xu, Jiahao and Liu, Qiuzhi and Chen, Xingyu and Wang, Yue and Song, Linfeng and Yu, Dian and Liang, Zhenwen and Wang, Wenxuan and others},
  journal={arXiv preprint arXiv:2504.11456},
  year={2025}
}

@misc{Qwen2.5-Math,
  author       = {{Qwen Team}},
  title        = {Qwen2.5-Math},
  year         = {2025},
  howpublished = {\url{https://github.com/QwenLM/Qwen2.5-Math}},
  note         = {Accessed: 2025-07-22}
}

@article{cui2025entropy,
  title={The entropy mechanism of reinforcement learning for reasoning language models},
  author={Cui, Ganqu and Zhang, Yuchen and Chen, Jiacheng and Yuan, Lifan and Wang, Zhi and Zuo, Yuxin and Li, Haozhan and Fan, Yuchen and Chen, Huayu and Chen, Weize and others},
  journal={arXiv preprint arXiv:2505.22617},
  year={2025}
}

@article{lou2025adacot,
  title={AdaCoT: Pareto-Optimal Adaptive Chain-of-Thought Triggering via Reinforcement Learning},
  author={Lou, Chenwei and Sun, Zewei and Liang, Xinnian and Qu, Meng and Shen, Wei and Wang, Wenqi and Li, Yuntao and Yang, Qingping and Wu, Shuangzhi},
  journal={arXiv preprint arXiv:2505.11896},
  year={2025}
}

@article{luo2025ada,
  title={Ada-R1: Hybrid-CoT via Bi-Level Adaptive Reasoning Optimization},
  author={Luo, Haotian and He, Haiying and Wang, Yibo and Yang, Jinluan and Liu, Rui and Tan, Naiqiang and Cao, Xiaochun and Tao, Dacheng and Shen, Li},
  journal={arXiv preprint arXiv:2504.21659},
  year={2025}
}

\appendix

\section{Algorithm}
Algorithm \ref{alg:ces} details the complete procedure for implementing Conditional Entropy Shaping (CES) within the DAPO framework.

\begin{algorithm*}[t]
\caption{CES within DAPO Framework}
\label{alg:ces}
\textbf{Input}: Prompt $x$, current policy $\pi_\theta$ \\
\textbf{Parameters}: Generation group size $N$, top-rate hyperparameter $\tau$, entropy scaling factors $\beta_1, \beta_2$ \\
\textbf{Output}: A set of shaped, token-level advantages $\mathcal{A}'$ for gradient update
\begin{algorithmic}[1] 
\STATE Generate response set $Y = \{y_1, \dots, y_N\}$ from $\pi_\theta( \cdot | x)$.
\STATE Compute rewards $R(y_i), r_{\text{acc}}(y_i)$ for each $y_i \in Y$.
\STATE Compute group accuracy $a$ based on $\{r_{\text{acc}}(y_i)\}$.
\STATE Initialize set of all shaped advantages $\mathcal{A}' \leftarrow \emptyset$.
\FOR{each response $y_i$ in $Y$}
    \STATE $A_i \leftarrow \text{GroupNormalize}(\{R(y_j)\}, R(y_i))$.
    \IF{$r_{\text{acc}}(y_i) = 1$}
        \STATE $b_i \leftarrow a$
    \ELSE
        \STATE $b_i \leftarrow 1 - a$
    \ENDIF
    \STATE Compute number of tokens to select $k_i = \lfloor |y_i| \cdot \tau \cdot b_i \rfloor$.
    \STATE $S_H(y_i) \leftarrow$ Identify top $k_i$ high-entropy tokens in $y_i$.
    \FOR{each token $t_j$ in $y_i$}
        \STATE $A'_{i,j} \leftarrow A_i$ \COMMENT{Initialize with base advantage}
        \IF{$t_j \in S_H(y_i)$}
            \STATE $H_j \leftarrow H(t_j | y_{i, <j})$  \COMMENT{Calculate entropy}
            \IF{$r_{\text{acc}}(y_i) = 1$}
                \STATE $A'_{i,j} \leftarrow A_i - \beta_1 \cdot H_j$ \COMMENT{Apply entropy penalty}
            \ELSE
                \STATE $A'_{i,j} \leftarrow A_i + \beta_2 \cdot H_j$ \COMMENT{Apply entropy reward}
            \ENDIF
        \ENDIF
        \STATE Add $A'_{i,j}$ to $\mathcal{A}'$.
    \ENDFOR
\ENDFOR
\STATE \textbf{return} $\mathcal{A}'$ \COMMENT{Return token-level shaped advantages $\mathcal{A}'$ for computing policy gradients in DAPO}
\end{algorithmic}
\end{algorithm*}

\section{Hyperparameters and Prompt}

\subsection{Training hyperparameters}
Table 3 lists the hyperparameters for our reinforcement learning experiments.

\begin{table}[t]
\centering
\label{tab:hyperparams}
\begin{tabular}{l c}
\toprule
\textbf{Hyperparameter} & \textbf{Value} \\
\midrule
\multicolumn{2}{c}{\textit{General Training Parameters}} \\
\midrule
Optimizer & Adam \\
Learning Rate & $2 \times 10^{-7}$ \\
Training Global Batch Size & 4 \\
Rollout Batch Size & 12 \\
Samples per Prompt & 4 \\
\midrule
\multicolumn{2}{c}{\textit{Generation Parameters}} \\
\midrule
Rollout Temperature & 1.0 \\
Rollout Top-$p$ & 1.0 \\
Maximum Generation Length & 12000 \\
\midrule
\multicolumn{2}{c}{\textit{CES-Specific Parameters}} \\
\midrule
Top-rate Hyperparameter ($\tau$) & 0.01 \\
Entropy Scaling Factor ($\beta_1, \beta_2$) & 0.4 \\
\bottomrule
\end{tabular}
\caption{Hyperparameters for RL training.}
\end{table}

\subsection{Evaluation prompt}
For all evaluation scenarios, we used the following standardized prompt to ensure the model generates answers in a step-by-step manner and formats the final result correctly:

\blockquote{
\textit{You are a helpful and harmless assistant. You should think step-by-step. 
Please put your final answer within \textup{\texttt{\textbackslash boxed\{\}}}.}
}

\section{Sensitivity to key hyperparameters}

To investigate the sensitivity of CES to its core hyperparameters and validate the robustness of CES, we conduct an ablation study on the top-rate $\tau$ and the entropy scaling factor $\beta_1, \beta_2$. We evaluate five different hyperparameter configurations on a representative subset of five datasets and compare their average performance against the DAPO baseline. The results are summarized in Table 4.

\begin{table}[ht]
\centering
\label{tab:hyper_ablation}
\begin{tabular}{cc cc}
\toprule
\textbf{$\boldsymbol{\tau}$} & \textbf{$\boldsymbol{\beta_1, \beta_2}$} & \textbf{Acc ↑} & \textbf{Len ↓} \\
\midrule
0.005 & 1.0 & 75.1 & 2855 \\
0.01  & 1.0 & 74.2 & 2992 \\
0.05  & 1.0 & 70.7 & 2818 \\
\midrule
\textbf{0.01} & \textbf{0.4} & \textbf{76.9} & \textbf{2757} \\
0.01  & 1.0 & 74.2 & 2992 \\
0.01  & 2.0 & 73.4 & 2997 \\
\midrule
\multicolumn{2}{l}{DAPO (Baseline)} & 72.6 & 3407 \\
\bottomrule
\end{tabular}
\caption{Hyperparameter sensitivity analysis for CES on the average of 5 datasets (AIME24, AMC23, GaoKao Math Cloze, GaoKao 2023 En and SVAMP). The optimal configuration is highlighted in \textbf{bold}.}
\end{table}

\subsection{\texorpdfstring{Analysis of Top-rate $\tau$}{Analysis of Top-rate tau}}
The hyperparameter $\tau$ controls the proportion of selected high-entropy tokens. With $\beta_1, \beta_2$ fixed at 1.0, we tested $\tau$ values of 0.005, 0.01, and 0.05. The results indicate that a smaller, more targeted intervention is more effective. As $\tau$ increases from 0.01 to 0.05, the average accuracy drops sharply from 74.2\% to 70.7\%, falling below the DAPO baseline. This suggests that selecting too many tokens introduces noise by including tokens that are not critical ``forking points'', thereby diluting the learning signal and degrading the policy.

\subsection{\texorpdfstring{Analysis of Entropy Scaling Factor $\beta_1, \beta_2$}{Analysis of Entropy Scaling Factor beta1, beta2}}
The hyperparameters $\beta_1, \beta_2$ indicate the scaling magnitude of the entropy reward and penalty. With $\tau$ fixed at 0.01, we tested $\beta_1, \beta_2$ values of 0.4, 1.0, and 2.0. The results show a clear trend: as $\beta_1, \beta_2$ increases, the average accuracy decreases while average response length increases. A larger setting on $\beta_1, \beta_2$ gives excessive weight to the entropy shaping term, particularly the exploratory reward on incorrect answers. This can cause the model to over-optimize for the process of exploration rather than the outcome of correctness, leading to longer, less focused reasoning chains that do not necessarily improve accuracy.

\subsection{Robustness of CES} 
Across four of the five tested hyperparameter settings, our method simultaneously outperforms the DAPO baseline in both accuracy and length. This demonstrates that CES provides consistent benefits across a reasonable range of hyperparameters, validating it as a stable and effective method for improving reasoning models.

\section{GPU Cost}

\begin{table}[ht]
\centering
\small
\setlength{\tabcolsep}{10pt}
\begin{tabular}{lccc}
\toprule
\textbf{Backbone} & \textbf{DAPO} & \textbf{CES} \\
\midrule
R1-7B   & \textbf{1.43 days} & 1.44 days \\
R1-1.5B & 9.36 hours         & \textbf{9.14 hours} \\
\bottomrule
\end{tabular}
\caption{Training GPU wall-clock time under the same setup.}
\label{tab:gpu_cost}
\end{table}

\begin{table*}[t!]
\centering
\small
\label{tab:results15b}
\begin{tabular}{*{7}{c}}
\toprule
\multirow{2}{*}{\textbf{Dataset}} & \multicolumn{2}{c}{\textbf{DAPO (Baseline)}} & \multicolumn{2}{c}{\textbf{CES (Ours)}} & \multicolumn{2}{c}{\textbf{Improvement}} \\
\cmidrule(lr){2-3} \cmidrule(lr){4-5} \cmidrule(lr){6-7}
& Acc $\uparrow$ & Len $\downarrow$ & Acc $\uparrow$ & Len $\downarrow$ & Acc $\uparrow$ & Len $\downarrow$ \\
\midrule
AIME24            & 6.7  & 10254 & \textbf{13.3} & \textbf{8751} & \textcolor{mygreen}{+6.6} & \textcolor{mygreen}{-1503} \\
AMC23             & 47.5 & 7291  & \textbf{55.0} & \textbf{6566} & \textcolor{mygreen}{+7.5} & \textcolor{mygreen}{-725} \\
CMATH             & 84.7 & \textbf{616}  & \textbf{85.3} & 676  & \textcolor{mygreen}{+0.6} & \textcolor{myred}{+60} \\
CN Middle School  & 59.4 & 794   & \textbf{61.4} & \textbf{758}  & \textcolor{mygreen}{+2.0} & \textcolor{mygreen}{-36} \\
College Math      & \textbf{34.1} & 2907  & 33.2 & \textbf{2866} & \textcolor{myred}{-0.9} & \textcolor{mygreen}{-41} \\
GaoKao Math Cloze & 46.6 & 5448  & \textbf{61.0} & \textbf{4655} & \textcolor{mygreen}{+14.4} & \textcolor{mygreen}{-793} \\
GaoKao 2023 En    & 57.7 & 3304  & \textbf{60.8} & \textbf{3173} & \textcolor{mygreen}{+3.1} & \textcolor{mygreen}{-131} \\
GSM8K             & 77.3 & \textbf{753}  & \textbf{78.6} & 1022 & \textcolor{mygreen}{+1.3} & \textcolor{myred}{+269} \\
Minerva Math      & 23.5 & 3418  & \textbf{29.0} & \textbf{3289} & \textcolor{mygreen}{+5.5} & \textcolor{mygreen}{-129} \\
Olympiad Bench    & 27.7 & 6868  & \textbf{32.0} & \textbf{6263} & \textcolor{mygreen}{+4.3} & \textcolor{mygreen}{-605} \\
SVAMP             & 85.6 & \textbf{618}  & \textbf{86.4} & 620  & \textcolor{mygreen}{+0.8} & \textcolor{myred}{+2} \\
TABMWP            & \textbf{79.2} & \textbf{699}  & 78.1 & 759  & \textcolor{myred}{-1.1} & \textcolor{myred}{+60} \\
\midrule
\textbf{Average}  & 52.5 & 3581 & \textbf{56.2} & \textbf{3283} & \textcolor{mygreen}{+3.7} & \textcolor{mygreen}{-298} \\
\bottomrule
\end{tabular}
\caption{Cross-scale generalization on DeepSeek-R1-Distill-1.5B. The best result in each category is in \textbf{bold}. ``Acc'' and ``Len'' denote the mean accuracy and the mean response length across 4 assessments for each benchmark.}
\end{table*}

A natural concern is whether CES introduces noticeable additional computation during training. Compared with vanilla DAPO, CES indeed adds two operations: token-level entropy computation and selection/shaping of high-entropy tokens. However, these additions do not require extra model forward or backward passes. In practice, entropy is computed directly from the logits that are already produced during rollout sampling, so the overhead is limited to lightweight vector reductions and top-k selection rather than additional Transformer backbone computation.

Results in Table \ref{tab:gpu_cost} indicates that the additional GPU time overhead of CES is negligible. For DeepSeek-R1-Distill-7B, DAPO takes 1.43 days, while CES takes 1.44 days, corresponding to only about +0.7\% relative overhead. For DeepSeek-R1-Distill-1.5B, DAPO takes 9.36 hours, while CES takes 9.14 hours, making CES slightly faster by about -2.3\%. These results suggest that the small constant-time overhead of entropy computation is largely offset by shorter rollouts during training.

\section{Generalization Experiments}

Although the main paper trains only on math data, the mechanism of CES is not inherently math-specific. It relies on token-level uncertainty (entropy) and correctness-conditioned shaping, which are \textbf{domain-agnostic signals}. We therefore evaluate generalization from two perspectives:

\begin{enumerate}
\item Across \textbf{model scales}.
\item Across \textbf{domains}.
\end{enumerate}

We find that the benefits of CES are not limited to the original 7B math setting, but generalize to a smaller backbone and to out-of-domain tasks in both general reasoning and code generation.

\subsection{Cross-Scale Generalization}
To test whether the benefits of CES depend on a single backbone scale, we additionally train DeepSeek-R1-Distill-1.5B under the same training protocol as the 7B model, and evaluate it on the same 12 math benchmarks.

Results in Table 6 show that CES remains effective on the 1.5B backbone. The average accuracy improves from 52.5 to 56.2, while the average response length is reduced from 3581 to 3283. Meanwhile, CES improves accuracy on most of the 12 benchmarks, and also shortens responses on the majority of them. Although a few benchmarks exhibit small length increases or minor accuracy fluctuations, the overall average still shows a clear win-win trend.

These results indicate that the benefit of CES is not tied to the 7B setting. When the backbone is scaled down to 1.5B, CES still consistently improves the accuracy?efficiency trade-off, suggesting that it is a generally useful training mechanism rather than a technique specific to a single model size.

\begin{table}[ht]
\centering
\small
\label{tab:general_reasoning}
\begin{tabular}{*{7}{c}}
\toprule
\multirow{2}{*}{\textbf{Benchmark}} & \multicolumn{2}{c}{\textbf{DAPO}} & \multicolumn{2}{c}{\textbf{CES}} \\
\cmidrule(lr){2-3} \cmidrule(lr){4-5}
& Acc $\uparrow$ & Len $\downarrow$ & Acc $\uparrow$ & Len $\downarrow$ \\
\midrule
ARC             & 79.7 & 543 & \textbf{83.6} & \textbf{542} \\
CommonsenseQA   & 56.5 & 451 & \textbf{65.8} & \textbf{448} \\
OpenBookQA      & 68.2 & \textbf{503} & \textbf{74.8} & 520 \\
\bottomrule
\end{tabular}
\caption{Cross-domain generalization on general reasoning benchmarks. The best result in each category is in \textbf{bold}.}
\end{table}

\subsection{Cross-Domain Generalization: General Reasoning}
To further test whether CES generalizes beyond the training distribution, we evaluate it on three general-reasoning benchmarks: ARC-Challenge, CommonsenseQA, and OpenBookQA. Importantly, these datasets are \textbf{outside the training domain}, since training uses only math data.

Results in Table 7 shows that the gains of CES do not simply come from forcing the model to generate shorter outputs. On ARC and CommonsenseQA, CES substantially improves accuracy while keeping the output length almost unchanged. On OpenBookQA, CES spends slightly more tokens in exchange for a meaningful gain in accuracy. In other words, CES does not learn a fixed preference for shorter responses; instead, it learns to allocate reasoning budget on demand. It permits additional reasoning when that helps correctness, and suppresses redundant exploration when it does not.

Therefore, these results suggest that the adaptive reasoning behavior learned by CES is not limited to math, but transfers to broader knowledge and commonsense reasoning tasks.

\subsection{Cross-Domain Generalization: Coding Benchmarks}

In addition to general reasoning, we also evaluate on code-generation benchmarks using EvalPlus. Specifically, we test on HumanEval / HumanEval+ and MBPP / MBPP+, where the ``+'' versions include stricter extra tests in addition to the original base tests.

\begin{table}[ht]
\centering
\small
\label{tab:coding_benchmarks}
\begin{tabular}{*{7}{c}}
\toprule
\multirow{2}{*}{\textbf{Benchmark}} & \multicolumn{2}{c}{\textbf{DAPO}} & \multicolumn{2}{c}{\textbf{CES}} \\
\cmidrule(lr){2-3} \cmidrule(lr){4-5}
& Acc $\uparrow$ & Len $\downarrow$ & Acc $\uparrow$ & Len $\downarrow$ \\
\midrule
HumanEval    & 46.3 & 4158 & \textbf{49.4} & \textbf{3911} \\
HumanEval+   & 43.9 & 4158 & \textbf{48.2} & \textbf{3911} \\
MBPP         & 42.9 & 2931 & \textbf{45.0} & \textbf{2824} \\
MBPP+        & 35.7 & 2931 & \textbf{39.4} & \textbf{2824} \\
\bottomrule
\end{tabular}
\caption{Cross-domain generalization on coding benchmarks. The best result in each category is in \textbf{bold}.}
\end{table}

Results are shown in Table 8, where CES outperforms DAPO on all four coding metrics. These results provide strong evidence of generalization, since code generation differs substantially from math reasoning in output format, structural constraints, and failure modes. Nevertheless, CES still improves both pass rate and token efficiency, suggesting that its optimization signal is task-agnostic. In addition, the improvements on HumanEval+ and MBPP+ indicate that the gains are robust under stricter extra tests, rather than appearing only on easier evaluation settings. Finally, the reduced generation length shows that CES does not improve coding results by ``thinking longer'', but by producing higher-quality solutions more efficiently.

Taken together, the coding results further support that CES generalizes beyond the math training domain to a structurally different reasoning task.

\section{Statistics of Simple vs. Hard Cases}

\begin{table}[ht]
\centering
\small
\setlength{\tabcolsep}{10pt}
\label{tab:simple_hard_cases}
\begin{tabular}{lccccc}
\toprule
\textbf{Dataset} & \textbf{Simple} & \textbf{Hard} & \textbf{Total} \\
\midrule
CMATH              & 534  & 66  & 600  \\
CN Middle School   & 60   & 41  & 101  \\
GSM8K              & 1121 & 198 & 1319 \\
TABMWP             & 870  & 130 & 1000 \\
\bottomrule
\end{tabular}
\caption{Statistics of simple and hard cases on several representative benchmarks. Following the definition in the main text, a question is classified as \emph{simple} if the original R1-7B achieves accuracy greater than 50\% on that question; otherwise, it is classified as \emph{hard}.}
\end{table}

Section 5.2 of the main paper notes that on a few relatively simple datasets, CES slightly increases the average response length. We report the numbers of simple and hard cases within these datasets in Table 9.

\end{document}